%% file: main.tex
\pdfoutput=1

\documentclass[11pt]{article}

\usepackage[preprint]{coling}

\usepackage{times}
\usepackage{latexsym}

\usepackage[T1]{fontenc}

\usepackage[utf8]{inputenc}

\usepackage{microtype}

\usepackage{inconsolata}

\usepackage{graphicx}
\usepackage{amsmath}
\usepackage{amssymb}

\usepackage{times}
\usepackage{latexsym}
\usepackage{graphicx}
\usepackage{wrapfig}
\usepackage{amsmath,bm}
\usepackage{amssymb}
\usepackage{algorithm}
\usepackage{algpseudocode}
\usepackage{multirow}
\usepackage{pifont}
\usepackage{wasysym}
\usepackage{amssymb}
\usepackage{color}
\usepackage{enumitem}
\usepackage{setspace}
\usepackage{indentfirst}
\usepackage{fnpct}
\usepackage{booktabs}

\newcommand{\x}{{\bf x}}
\newcommand{\p}{{\bf p}}

\newcommand{\z}{{\bf z}}
\newcommand{\X}{\text{X}}

\newcommand{\btheta}{\bm{\theta}}
\newcommand{\bphi}{\bm{\phi}}
\newcommand{\EE}{{\mathbb{E}}}
\newcommand{\norm}[1]{\left\lVert#1\right\rVert}

\title{VaeDiff-DocRE: End-to-end Data Augmentation Framework \\ for Document-level Relation Extraction}

\begin{document}
\author{
     \textbf{Khai Phan Tran\textsuperscript{$\spadesuit$}},
     \textbf{Wen Hua\textsuperscript{$\clubsuit$}},
     \textbf{Xue Li\textsuperscript{$\spadesuit$}}
    \\
    \\
     \normalsize \textsuperscript{$\spadesuit$} School of Electrical Engineering and Computer Science, The University of Queensland, Australia \\
     \normalsize \textsuperscript{$\clubsuit$} Department of Data Science and Artificial Intelligence, The Hong Kong Polytechnic University, \\ Hong Kong SAR, China
    \\
     \small{
       \textbf{Correspondence:} \href{mailto:phankhai.tran@uq.edu.au}{phankhai.tran@uq.edu.au}
     }
}
\maketitle
\begin{abstract}
    Document-level Relation Extraction (DocRE) aims to identify relationships between entity pairs within a document. However, most existing methods assume a uniform label distribution, resulting in suboptimal performance on real-world, imbalanced datasets. To tackle this challenge, we propose a novel data augmentation approach using generative models to enhance data from the embedding space. Our method leverages the Variational Autoencoder (VAE) architecture to capture all relation-wise distributions formed by entity pair representations and augment data for underrepresented relations. To better capture the multi-label nature of DocRE, we parameterize the VAE's latent space with a Diffusion Model. Additionally, we introduce a hierarchical training framework to integrate the proposed VAE-based augmentation module into DocRE systems. Experiments on two benchmark datasets demonstrate that our method outperforms state-of-the-art models, effectively addressing the long-tail distribution problem in DocRE. Our code is released at: {\tt \href{https://github.com/khaitran22/VaeDiff-DocRE}{https://github.com/khaitran22/VaeDiff \\-DocRE}}
\end{abstract}

\input{sections/intro}
\input{sections/method}
\input{sections/experiments}
\input{sections/related_work}
\input{sections/conclusion}
\input{sections/limitation}
\input{sections/acknowledgements}

\bibliography{bib}
\appendix
\input{sections/appendix}

\end{document}

%% file: sections/intro.tex
\section{Introduction}
\noindent
The task of relation extraction (RE) involves identifying possible relationships between two entities within a given context. This is a fundamental task in the field of Information Extraction, as it serves as a foundation for other tasks such as Question Answering \cite{xu2016question}. Early approaches to RE focused on sentence-level settings, where the context is limited to a single sentence. However, this approach is often impractical, as many relationships can only be inferred from information spread across multiple sentences. For instance, at least 40.7\% of relational facts on Wikipedia can only be identified from multiple sentences \cite{yao2019docred}. As a result, there has been growing interest in exploring the problem of extracting relations across multiple sentences, a task known as Document-level Relation Extraction (DocRE) \cite{xu2021document, zhang2021document, zhou2021document, tan2022document}.

\begin{figure}[t]
    \centering
    \includegraphics[width=0.93\linewidth]{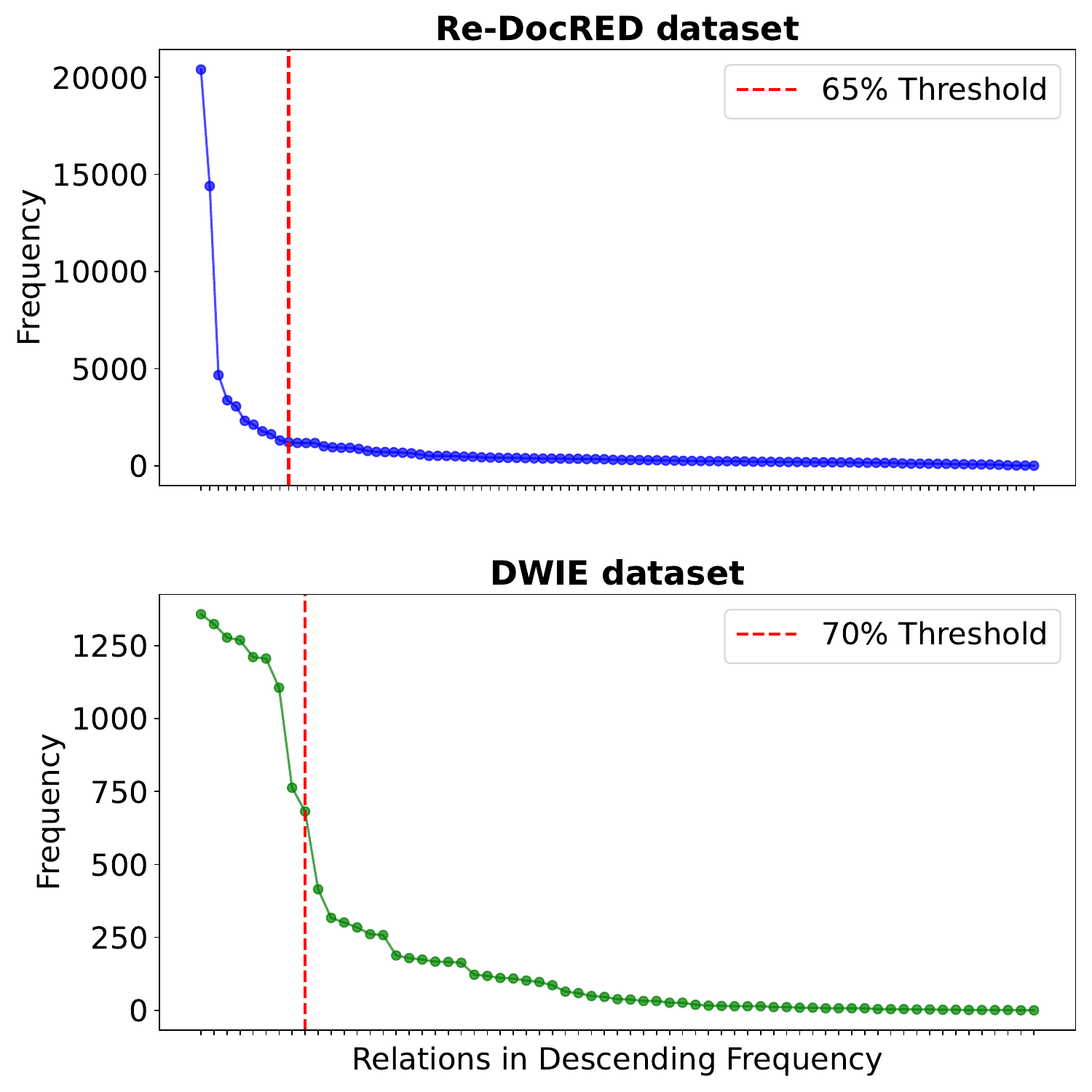}
    \caption{Relation frequency in Re-DocRED \cite{tan2022revisiting} and DWIE \cite{zaporojets2021dwie} datasets.}
    \label{fig:dist-label}
\end{figure}

Notably, the distribution of relations in datasets for the DocRE task often exhibits significant imbalance, where a few relation labels dominate the majority of instances. This imbalance is apparent both in the disparity between entity pairs that express at least one positive relation and those that do not, a situation we refer to as \textit{positive-negative imbalance} (PN imbalance), and among the relation labels themselves, which we call \textit{positive-positive imbalance} (PP imbalance). For example, in the DocRED dataset \cite{yao2019docred}, at least 94\% of entity pairs do not express any relation (PN imbalance). Additionally, 10 out of 96 relation labels account for more than 60\% of the training instances (PP imbalance), while in the DWIE dataset \cite{zaporojets2021dwie}, 9 out of 66 relation labels account for more than 70\% of the training instances, as illustrated in Figure \ref{fig:dist-label}. Consequently, training DocRE systems on these datasets often leads to degraded performance due to the skewed distribution.

Recent research in the DocRE task has concentrated on methods to address distribution imbalance \cite{zhang2021document, zhou2021document, tan2022document, zhou2022none, guo2023towards, wang2023adaptive}. However, these studies primarily focus on mitigating the PN imbalance, often overlooking the PP imbalance phenomenon. Only a few works, notably \citet{tan2022document} and \citet{guo2023towards}, have specifically investigated the PP imbalance issue. They proposed different objective functions aimed at enhancing the distinction between entity pair representations across different relation labels. However, these approaches tend to improve the performance of one type of class at the expense of another \cite{zhang2023deep}. Empirically, we observe that the objective function in \citet{tan2022document} enhances performance on minority classes but at the cost of majority classes. Additionally, removing the modified version of supervised contrastive loss proposed by \citet{guo2023towards} — intended to address the PP imbalance issues — can paradoxically lead to better overall performance. A feasible solution to addressing the trade-off and PP imbalance is to increase the number of training instances for minority classes.

\begin{figure}[ht]
    \centering
    \includegraphics[width=\columnwidth]{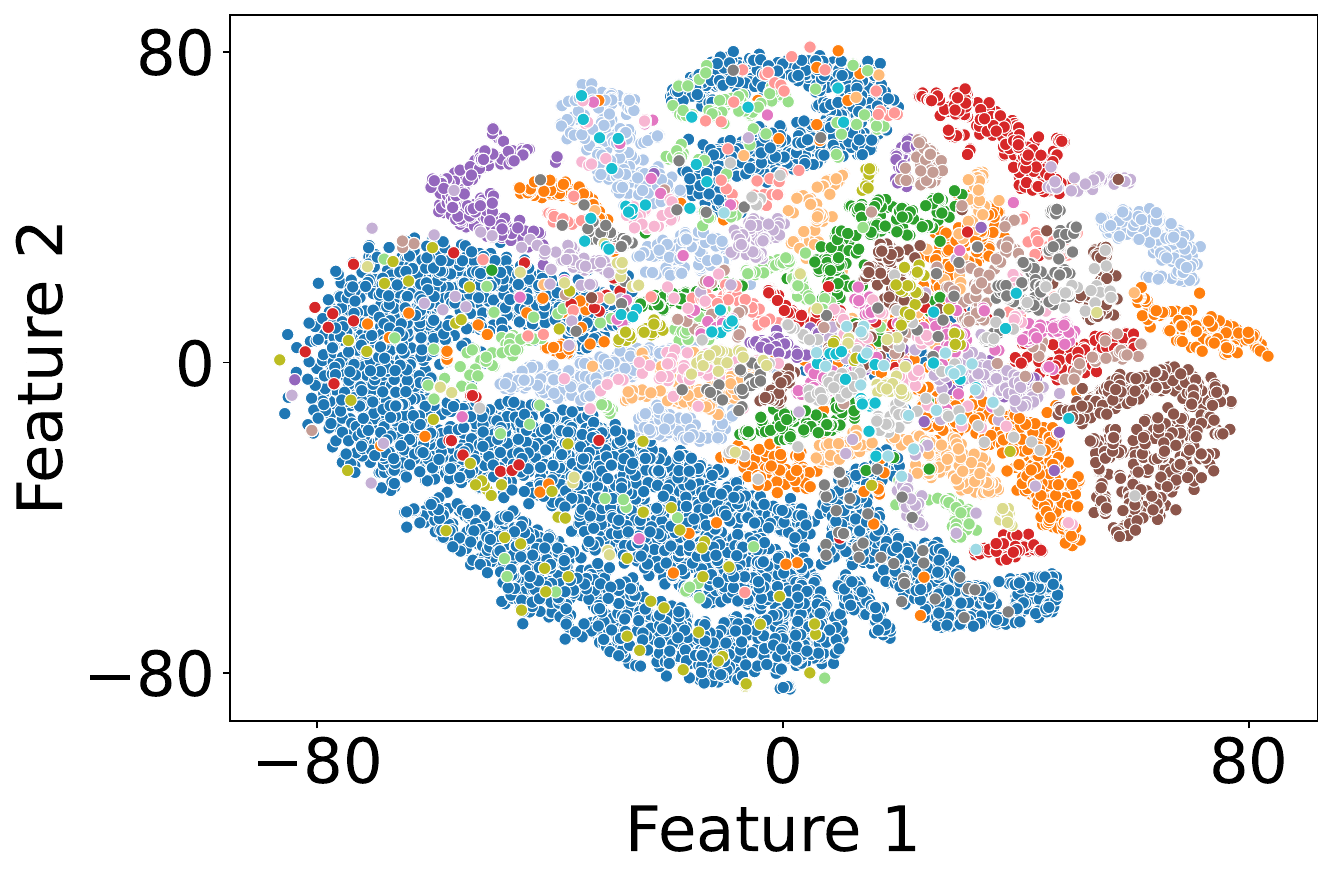}
    \caption{Relation-wise distribution from Re-DocRED dataset visualized by t-SNE \cite{van2008visualizing}.}
    \label{fig:structured-pair-emb}
\end{figure}

Inspired by this, we propose a novel data augmentation approach in the DocRE task. Since preprocessing and labeling datasets for the DocRE task are resource-intensive and prone to human errors (e.g., false negatives), we propose a data augmentation technique in the embedding space for DocRE. Specifically, we observe that after the training process, the representation of entity pairs forms distinct relation-wise clusters, as shown in Figure \ref{fig:structured-pair-emb}. This observation suggests that each relation has its own underlying distribution, shaped by the embeddings of entity pairs expressing that relation. By learning these distributions, we can generate entity pair representations that accurately convey the targeted relations within the embedding space.

Building on this observation, we propose a deep generative model approach to augment data for the DocRE task. This approach integrates a Variational Auto-Encoder (VAE) \cite{kingma2013auto} with a Diffusion Probabilistic Model (DPM) \cite{ho2020denoising}. Assuming that each relation exhibits a distinct distribution shaped by the entity pair representations expressing that relation, as illustrated in Figure \ref{fig:structured-pair-emb}, we train the VAE to capture these relation-wise distributions in the embedding space. The DPM architecture is then employed to parameterize the latent space of the VAE, effectively addressing the multi-label nature of the DocRE task. We call this module as VaeDiff. Finally, we introduce a hierarchical training framework that learns the relation-wise distributions, trains VaeDiff, and integrates it to improve the performance of the DocRE model, particularly under long-tail distribution settings.

We name our overall framework as \textbf{\underline{V}}ariational \textbf{\underline{A}}uto\textbf{\underline{e}}ncoder and \textbf{\underline{Diff}}usion Prior for \textbf{\underline{Doc}}ument-level \textbf{\underline{R}}elation \textbf{\underline{E}}xtraction (VaeDiff-DocRE). We conduct the experiments on two DocRE dataset - the Re-DocRED \cite{tan2022revisiting} and the DWIE \cite{zaporojets2021dwie}, which demonstrate that our VaeDiff-DocRE model significantly outperforms the state-of-the-art methods in addressing long-tail issues in DocRE task. The contribution of our work are summarized as follows:
\begin{itemize}
    \item We propose a data augmentation framework via embedding space to address the PP imbalance and the performance trade-off issues of existing methods.
    \item We propose leveraging VAE and DPM architectures in our framework for augmenting data. To the best of our knowledge, we are the first applying DPM to the DocRE task.
    \item Extensive experiments on two public DocRE datasets, Re-DocRED \cite{tan2022revisiting} and DWIE \cite{zaporojets2021dwie}, demonstrate the effectiveness of our VaeDiff-DocRE framework compared to state-of-the-art baselines.
\end{itemize}

%% file: sections/method.tex
\section{Methodology}
\noindent
In this section, we introduce our VaeDiff-DocRE framework, a novel data augmentation approach at the embedding space to assist the DocRE task. Section \ref{sec-vaediff} introduces the data augmentation module VaeDiff, which comprises a VAE architecture to model relation-wise distributions in the embedding space and a DPM to parameterize the prior distribution of the VAE, specifically designed for multi-label settings. Section \ref{sec-hier-train-appr} presents VaeDiff-DocRE, an end-to-end training framework that learns the relation-wise distributions, trains VaeDiff, and integrates it to enhance the performance of the DocRE model.

\subsection{Problem Formulation}
\noindent
Let $D\!=\!\{w_l\}_{l=1}^{L}$ be a document containing $L$ words and a set of entities $\mathcal{E}_D = \{e_i\}_{i=1}^{|\mathcal{E}_D|}$. Each entity $e_i$ is associated with a set of mentions $\mathcal{M}_{e_i}\!=\!\{m^i_j\}_{j=1}^{|\mathcal{M}_{e_i}|}$ (i.e., a set of phrases referring to the same entity $e_i$). In DocRE task, we predict the subset of relations in a predefined set $\mathcal{R}\!=\!\{r_k\}_{k=1}^{|\mathcal{R}|}$ that hold between each pair of entities $(e_h, e_t)_{h,t=1,\ldots,|\mathcal{E}_D|, h \neq t}$. We sometimes abbreviate an entity pair $(e_h,e_t)$ as $(h,t)$ to simplify notation. A relation is deemed to exist between the head entity $e_h$ and tail entity $e_t$ if it is expressed between any of their corresponding mentions. If no relation exists between any pair of their mentions, the entity pair is labeled {\tt NA}. For each entity pair, we term a relation that holds between its constituent entities as \textit{positive}, and the remaining relations in $\mathcal{R}$ as \textit{negative}. An entity pair that is {\tt NA} does not have any positive relation, and has the entire set $\mathcal{R}$ as negative relations (we could consider such a pair as having a special {\tt NA} relation between them).

\subsection{Data Augmentation Module - VaeDiff}
\label{sec-vaediff}
\noindent
The proposed VaeDiff module consists of a VAE architecture, called the Entity Pair VAE (EP-VAE), designed to learn relation-wise distributions with a compact latent space. We then parameterize the latent space in EP-VAE using a diffusion model, referred to as the Diffusion Prior, specifically tailored for multi-label settings in DocRE. During inference, the Diffusion Prior generates relation-specific latent variables, which are subsequently decoded by the EP-VAE to generate new entity pair representations corresponding to the target relation.

\subsubsection{Entity Pair VAE}
\label{sec:ep-vae}
\begin{figure}[!t]
    \centering
    \includegraphics[width=\columnwidth]{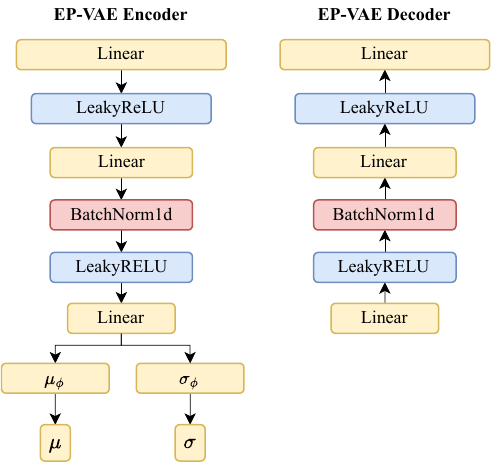}
    \caption{Structure of EP-VAE module.}
    \label{fig:ep-vae-architecture}
\end{figure}

\noindent
The Entity Pair VAE, referred to as EP-VAE, consists of an encoder and a decoder $f_{\btheta}(\cdot)$, parameterized by \( \btheta \). The encoder includes two components: (1) a projector \( q_{\bphi}(\cdot) \) that maps the input to the EP-VAE latent space, and (2) separate projectors for the mean (\( \mu_\phi \)) and log variance (\( \sigma_\phi \)), which estimate the parameters of the posterior distribution in the latent space. Both the encoder projector and the decoder share an identical architecture comprising alternating feedforward layers and LeakyReLU activation functions, with a single BatchNorm layer \cite{ioffe2015batch} incorporated mid-stack. The mean (\( \mu_\phi \)) and log variance (\( \sigma_\phi \)) projectors are implemented as feed-forward networks. The detailed structure of EP-VAE is depicted in Figure \ref{fig:ep-vae-architecture}.

\begin{figure*}[!ht]
    \centering
    \includegraphics[width=\linewidth]{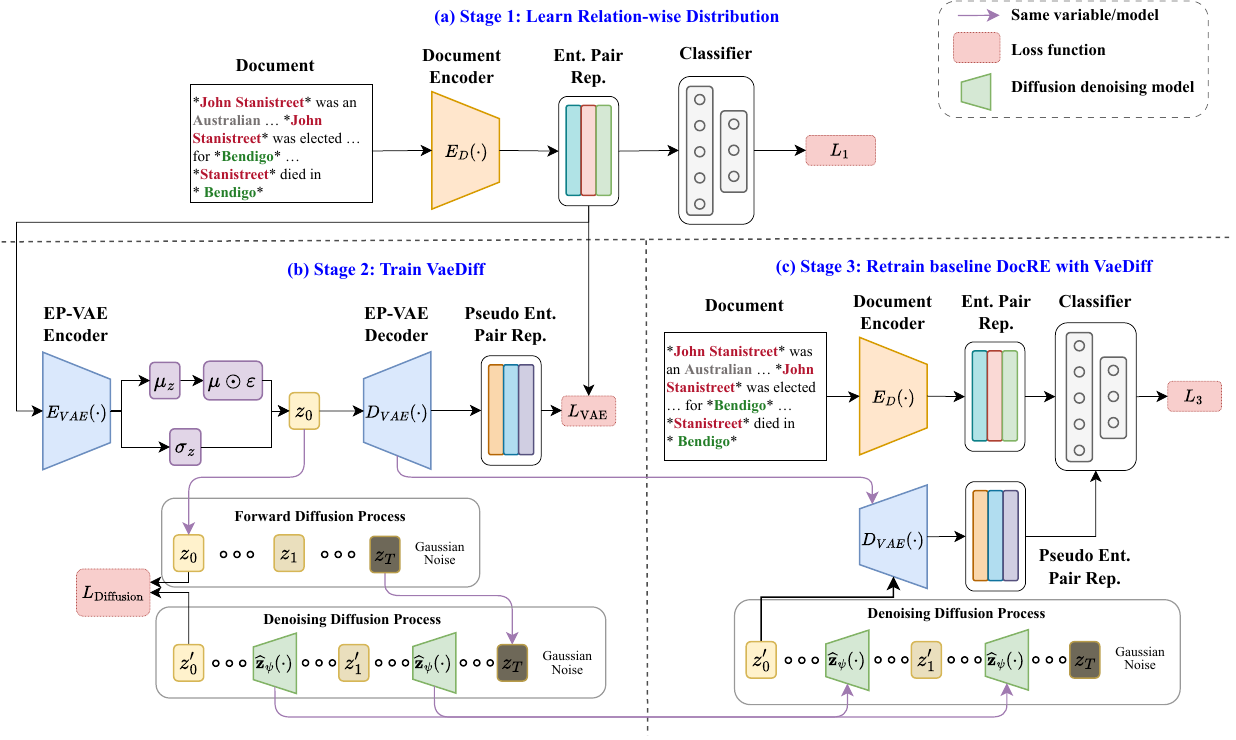}
    \caption{Overview of our VaeDiff-DocRE framework.}
    \label{fig:vaediff-framework}
\end{figure*}

Let \(\mathcal{P} = (\p_1, \p_2, \dots, \p_n)\) represent the set of training entity pair representations, and let \(\z_p\) denote the latent vectors. The projector \(q_{\bphi}(\cdot)\) first maps the input \(\p\) to a latent space representation \(\p'\) of the EP-VAE framework. Subsequently, the mean (\(\mu_\phi\)) and log variance (\(\sigma_\phi\)) projectors produce the parameters \(\boldsymbol{\mu}_p\) and \(\log (\boldsymbol{\sigma}_p)\), which define the posterior distribution \(q_\phi(\z_p | \p)\):
\begin{equation}
    \begin{aligned}
        \p' &= q_{\bphi}(\p) \\
        \boldsymbol\mu_p &= \mu_\phi(\p') \\
        \log (\boldsymbol\sigma_p) &= \sigma_\phi(\p')
    \end{aligned}
\end{equation}
where $\boldsymbol{\mu_p}$ and $\boldsymbol\sigma_p$ are parameters of multivariate Gaussian distribution. To enable back-propagation, we use reparameterization trick proposed by \citet{kingma2013auto} as follows:
\begin{equation}
    \label{reparam-vae}
    \mathbf{z}_p = \bm{\mu}_p + \bm{\sigma}_p \odot \bm{\epsilon}, \; \text{where} \; \bm{\epsilon} \sim \mathcal{N}(\bm{0}, \boldsymbol{I}).
\end{equation}

Then, $\mathbf{z}_p$ are given to the decoder $f_{\btheta}(\cdot)$ to reconstruct the original entity pair representation as follows:
\begin{equation}
    \Tilde{\p} = f_{\btheta}(\mathbf{z}_p)
\end{equation}

Finally, the EP-VAE is trained to minimize the negative Evidence Lower BOund (ELBO) \cite{kingma2013auto} as Eq. (\ref{eq:elbo}):
\begin{equation} 
    \label{eq:elbo}
    \resizebox{.86\linewidth}{!}{
        $\begin{aligned}
            \mathcal{L}_\text{VAE} &= -\sum_{i=1}^n \EE_{q_{\bphi}(\z|\p)} \left[\log(f_{\btheta}(\p_i|\mathbf{z}_{\p_i}))\right] \\ 
            & - D_\text{KL}\left[q_{\bphi}(\mathbf{z}_{\p_i}|\p_i) || p(\mathbf{z}_{\p_i})\right]
        \end{aligned}$
    }
\end{equation}
\noindent
where $p(\z_\p)$ is the prior distribution of the latent variable $\z_\p$. 

The first term is the reconstruction loss, corresponding to the cross-entropy between the original entity pair representation $\p_i$ and its reconstruction $\Tilde{\p}_i$. This term ensures the accurate reconstruction of each entity pair representation $\p_i \in \mathcal{P}$. The second term is the Kullback-Leibler divergence ($D_\text{KL}$) between the prior distribution of the latent vectors $p(\mathbf{z}_\p)$ and the posterior distribution $q_{\bphi}(\z_\p|\p)$. This term acts as a regularization factor, encouraging the posterior distribution of the entity pair representation to remain close to the prior.

\subsubsection{Diffusion Prior}
\noindent
In the original VAE formulation, the prior distribution \( p(\mathbf{z}_\p) \) is assumed to follow a standard normal distribution, which limits the latent space's ability to model the context in a multi-label scenario. This can reduce the effectiveness of the generated entity pair representations for target relations. To address this, we introduce a Diffusion Prior by parameterizing \( p(\mathbf{z}_\p) \) with a DPM, enhancing the complexity of the latent space in the EP-VAE. This new prior is denoted as \( p_{\psi}(\mathbf{z}_\p) \), where \( \psi \) represents the model's parameters. Additional background on DPM can be found in Appendix \ref{ddpm}.

\paragraph{Architecture.} Our denoising network $\hat{\z}_\psi(\cdot)$ is a pre-LayerNorm transformer \cite{xiong2020layer}. Each layer is conditioned on the timestep $t$ by incorporating the Transformer sinusoidal position embedding \cite{vaswani2017attention}, followed by an MLP with a single hidden layer.

\paragraph{Objective Function.}
Unlike the approach proposed by \citet{ho2020denoising}, we discovered that training our denoising network $\hat{\z}_\psi(\z_t, t)$ to directly predict the original latent vectors $\Tilde{\z}_0$ from its arbitrarily noisified version $\z_t$ yields better results. Thus, we train our Diffusion Prior with the objective function as below:
\begin{equation}
    \label{eq:diff-loss}
    \mathcal{L}_\text{Diff} = \norm{\hat{\z}_\psi(\z_t, t) - \z_0}^2_2
\end{equation}

\paragraph{Self-condition} To improve denoising performance, we implement self-conditioning technique introduced by \citet{chen2023analog}. Normally, the network is conditioned on the latent variable and timestep: \(\Tilde{\z}^t_0 = \hat{\z}_\psi(\z_t, t)\). Self-conditioning adds an estimate from the previous timestep \(s\): \(\Tilde{\z}^t_0 = \hat{\z}_\psi(\z_t, t, \Tilde{\z}^s_0)\), enhancing the current prediction.

During training, the network alternates between without \(\Tilde{\z}^{t, \emptyset}_0 = \hat{\z}_\psi(\z_t, t, \emptyset)\) and with self-conditioning by probability \(p\), where \(\Tilde{\z}^{t, \emptyset}_0 = \hat{\z}_\psi(\z_t, t, \emptyset)\) is computed first and refined to \(\Tilde{\z}^t_0 = \hat{\z}_\psi(\z_t, t, \text{sg}(\Tilde{\z}^{t, \emptyset}_0))\) with $\text{sg}(\cdot)$ denotes the stop-gradient operation. A learnable embedding is used when no previous estimate is available, like in the without self-conditioning.

\paragraph{Class Condition} For multi-label settings, we propose leveraging a conditional diffusion model using the Classifier-Free Diffusion Guidance paradigm \cite{ho2021classifierfree}. The model alternates between conditional generation \(\Tilde{\z}^{t}_0 = \hat{\z}_\psi(\z_t, t, y)\) and unconditional generation \(\Tilde{\z}^{t, \emptyset}_0 = \hat{\z}_\psi(\z_t, t, \emptyset)\), with probability \(p\). The inference output \(\Tilde{\z}^t_0\) is a weighted combination of conditional and unconditional outputs with a hyper-parameter $w$:

\begin{equation}
    \label{eq:diff-inf}
    \Tilde{\z}^t_0 = (1 - w) \cdot \Tilde{\z}^{t, y}_0 + w \cdot \Tilde{\z}^{t, \emptyset}_0
\end{equation}

We introduce a learnable embedding \(\mathbf{y} \in \mathbb{R}^{|\mathcal{R}| \times d}\) for each label, and for multi-label cases, the conditional input is the sum of embeddings for all ground truth labels:

\begin{equation}
    y = \sum_{i \in \mathcal{R}} \mathbb{I}(y_i = 1) \cdot \mathbf{y}_i
\end{equation}

We also incorporate conditional information into the time embedding.

\subsubsection{Training Objective}
\noindent
To co-train Diffusion Prior and EP-VAE, we decompose the Kullback-Leibler divergence term \( D_\text{KL}\) in Eq. (\ref{eq:elbo}) into its negative encoder entropy and cross-entropy. Specifically, the decomposition is as follows:
\begin{equation}
    \label{eq:elbo-decomp}
    \resizebox{.89\linewidth}{!}{
        $\begin{aligned}
            D_\text{KL}\left[q_{\bphi}(\mathbf{z}_{\p}|\p) || p_{\psi}(\mathbf{z}_{\p})\right] &= \underbrace{\mathbb{E}_{q_{\bphi}(\mathbf{z}_{\p}|\p)}\left[\log(q_{\bphi}(\mathbf{z}_{\p}|\p))\right]}_\text{negative encoder entropy} \\ 
            & + \underbrace{\mathbb{E}_{q_{\bphi}(\mathbf{z}_{\p}|\p)}\left[-\log(p_{\psi}(\mathbf{z}_{\p}))\right]}_\text{cross-entropy}
        \end{aligned}$
    }
\end{equation}
Following \citet{wehenkel2021diffusion}, the cross-entropy term can further be replaced by the regression objective \( \mathcal{L}_\text{Diff} \) of the diffusion model in Eq. (\ref{eq:diff-loss}). As a result, the VaeDiff component is trained with the following objective function:

\begin{equation}
    \label{eq:loss-vaediff}
    \resizebox{.8\linewidth}{!}{
        $\begin{aligned}
            \mathcal{L}_\text{VaeDiff} &= \underbrace{\mathbb{E}_{q_{\bphi}(\mathbf{z}_\p|\p)} \left[-\log(f_{\btheta}(\p|\mathbf{z}_{\p}))\right]}_\text{reconstruction EP-VAE loss} \\ 
            & + \underbrace{\mathbb{E}_{q_{\bphi}(\mathbf{z}_\p|\p)} \left[\log(q_{\bphi}(\mathbf{z}_{\p}|\p))\right]}_\text{negative EP-VAE encoder entropy} \\ 
            & + \underbrace{\mathbb{E}_{q_{\bphi}(\mathbf{z}_{\p}|\p)}\left[\norm{\hat{\mathbf{z}}_\psi(\mathbf{z}_t, t) - \mathbf{z}_0}^2_2\right]}_{\mathcal{L}_\text{Diff}}
        \end{aligned}$
    }
\end{equation}
The detailed derivation of Eq. (\ref{eq:loss-vaediff}) is presented in Appendix \ref{sec:derive-eq-loss-vaediff}.

\subsection{Hierarchical Training Data Augmentation Framework - \textbf{VaeDiff-DocRE}}
\label{sec-hier-train-appr}
\noindent
In this section, we introduce VaeDiff-DocRE, an end-to-end hierarchical three-stage training data augmentation framework designed to improve the performance of the DocRE system under long-tail distribution settings. The overall framework is depicted in Figure \ref{fig:vaediff-framework}.

\subsubsection{Stage 1: Learn Relation-wise Distribution} \label{stage-1}
\noindent
To obtain the structured embedding space depicted in Figure \ref{fig:structured-pair-emb}, we train a simple baseline DocRE model, which initially performs lower than the compared baseline methods. The primary goal of this baseline is to mitigate the negative impact of the PN imbalance, which is not the main focus of this work, and to demonstrate the effectiveness of our proposed data augmentation approach. This baseline DocRE model can be replaced by any more advanced DocRE model to achieve better results. For entity pair representation learning, we adopt the structure of KD-DocRE \cite{tan2022document} as the backbone for our baseline model. Regarding the objective function, we replace the Adaptive Focal Loss used in KD-DocRE with PMTEM loss \cite{guo2023towards} and Supervised Contrastive Learning loss \cite{khosla2020supervised}. The detailed structure of this baseline model is provided in Appendix \ref{baseline-docre}.

\subsubsection{Stage 2: Train VaeDiff Module}
\noindent
Following the training of the baseline DocRE model in stage 1, we proceed to train the VaeDiff augmentation module as described in Section \ref{sec-vaediff}. The training inputs for the VaeDiff module are generated by utilizing the document encoder from the trained DocRE baseline to extract entity pair representations within a document. Initially, the EP-VAE is trained independently, without co-training the Diffusion Prior, to allow the latent space to stabilize over the first \( n \) epochs. Subsequently, the EP-VAE and the Diffusion Prior are co-trained using the objective function specified in Eq. (\ref{eq:loss-vaediff}).

\subsubsection{Stage 3: Retrain baseline DocRE with VaeDiff}
\noindent
In the final stage, we retrain the baseline DocRE model using the trained VaeDiff augmentation model to address the PP imbalance in the DocRE task. The detailed algorithm of stage 3 is provided in Algorithm \ref{alg:stage-3}.

Following the warmup approach in stage 2, we initially train the baseline DocRE model for the first $n$ epochs without augmenting any new entity representations. After training for initial $n$ epochs, we then train the DocRE model using both the features $\mathbf{x}$ from its document encoder and the pseudo-features $\Tilde{\mathbf{x}}$ generated by the VaeDiff model.

To generate $\Tilde{\mathbf{x}}$, assume there are $|\mathcal{E}_p|$ positive entity pairs in the current batch $\mathcal{B}$. We select the label vectors $l_{\mathcal{E}_p}$ of these positive entity pairs. The VaeDiff model then generates $m \times |\mathcal{E}_p|$ additional entity pair representations using $l_{\mathcal{E}_p}$ as class conditional information. The DocRE model makes predictions based on the combined set $\mathbf{X} = \{\mathbf{x}, \Tilde{\mathbf{x}}\}$, and these predictions are concatenated to compute the loss for training the DocRE model.

\begin{algorithm}[ht]
    \label{stage-3-appendix}
    \footnotesize
    \textbf{Input:} Dataset $\mathcal{D}$ \\
    \textbf{Model:} DocRE model $f_\xi(\cdot)$, Diffusion Prior model $\hat{\mathbf{z}}_\psi(\cdot)$, EP-VAE Decoder $f_\theta(\cdot)$
    \caption{{\tt VaeDiff-DocRE} framework}
        \label{alg:stage-3}
        \begin{algorithmic}[1]
            \While{epoch $\leq$ total\_epochs}
            \State $(\mathbf{x}, \mathbf{y}) \sim \mathcal{D}$ \Comment{Sample batch data}
            \If{epoch $\leq$ warmup\_epochs} \label{start-w}
                \State $\hat{\mathbf{y}} \gets f_\xi(\mathbf{x})$ \Comment{Predict relation labels}
                \State Compute $\mathcal{L}(\hat{\mathbf{y}}, \mathbf{y})$ by Eq (\ref{base-loss}). \label{end-w} \\
            \Else \Comment{Ent. Pair Rep Augmentation Process}
                \State $\mathbf{z}_T \sim \mathcal{N}(0, \mathbf{I})$ \Comment{Sample noise}    \label{start-aug}
                \For{$t = T, ..., 1$} \Comment{Denoising process} \\
                    \State {\tt // Self-conditioning}
                    \If{$t == T$}
                        \State $\Tilde{\z}^{t, y}_0 \gets \hat{\z}_\psi(\z_t, t, \emptyset, \mathbf{y})$
                        \State $\Tilde{\z}^{t, \emptyset}_0 \gets \hat{\z}_\psi(\z_t, t, \emptyset, \emptyset)$
                    \Else
                        \State $\Tilde{\z}^{t, y}_0 \gets \hat{\z}_\psi(\z_t, t, \text{sg}(\Tilde{\z}^{t+1}_0), \mathbf{y})$
                        \State $\Tilde{\z}^{t, \emptyset}_0 \gets \hat{\z}_\psi(\z_t, t, \text{sg}(\Tilde{\z}^{t+1}_0), \emptyset)$
                    \EndIf
                    \\
                    \State {\tt // Class-conditioning}
                    \State $\Tilde{\z}^t_0 = (1 - w) \cdot \Tilde{\z}^{t, y}_0 + w \cdot \Tilde{\z}^{t, \emptyset}_0$ \\
                    \State {\tt // Sample sampling}
                    \State $\Tilde{\bm{\epsilon}} \sim \mathcal{N}(0, \mathbf{I})$ if $t > 1$ else $\Tilde{\bm{\epsilon}} = 0$
                    \State Compute $\tilde{\bm{\mu}}_t$ as in DDPM sampling
                    \State $\mathbf{z}_{t-1} = \tilde{\bm{\mu}}_t + \sigma_t * \Tilde{\bm{\epsilon}}$ \\
                \EndFor
                \State $\Tilde{\mathbf{x}} \gets f_\theta(\Tilde{\z}_0)$ \Comment{Generating New Rep} \label{end-aug} \\

                \State {\tt // Prediction}
                \State $\hat{\mathbf{y}}, \Tilde{\mathbf{y}} \gets f_\xi(\mathbf{x}), f_\xi(\Tilde{\mathbf{x}})$ \label{start-pred}
                \State $\hat{\mathbf{y}}' = {\tt Concat}([\Tilde{\mathbf{y}}, \hat{\mathbf{y}}])$
                \State Compute $\mathcal{L}(\hat{\mathbf{y}}', \mathbf{y})$ by Eq (\ref{base-loss}). \label{end-pred}
            \EndIf
            \State Update DocRE model $f_\xi(\cdot)$
            \EndWhile
        \end{algorithmic}
\end{algorithm}

%% file: sections/experiments.tex
\section{Experiments}
\begin{table*}[!t]
    \centering
    \resizebox{\textwidth}{!}{%
    \begin{tabular}{@{}lccccccccc@{}}
        \toprule
        \multirow{2}{*}{\textbf{Method}}                                                                              & \multirow{2}{*}{\textbf{PLM}} & \multicolumn{4}{c}{\textbf{Dev}}                                                                                  & \multicolumn{4}{c}{\textbf{Test}}                                                                                 \\ \cmidrule(l){3-6}  \cmidrule(l){7-10}
                                                                                                                      &                               & Ign F1 & F1    & Freq F1 & LTail F1 & Ign F1 & F1    & Freq F1 & LTail F1 \\ \midrule
        \textbf{LLM-based}      &  &  &  &  &  &  & &  &   \\
        GPT-3.5 + NLI \cite{li2023semi}      & GPT-3.5 & - & - & - & - & - & 9.33 & - & -  \\
        ZeroDocRTE \cite{sun2024consistency}    & LLaMA2-13B-Chat & - & 11.90 & - & - & - & 13.23 & - & -                                  \\
        ZeroDocRE \cite{sun2024consistency}     & LLaMA2-13B-Chat & - & 48.51 & - & - & - & 49.21 & - & -                                  \\
        AutoRE \cite{xue2024autore}          & Vicuna-7B & - & 54.29 & - & - & - & 53.84 & - & -                                  \\
        LMRC \cite{li2024llm}                & LLaMA2-13B-Chat & - & - & - & - & 74.08 & 74.63 & - & -                                  \\\midrule
        CNN \cite{yao2019docred}\textsuperscript{\textdagger}                                                                     & -                             & 53.95                            & 55.60 & -                                 & -                                  & 52.80                            & 54.88 & -                                 & -                                  \\
        LSTM \cite{yao2019docred}\textsuperscript{\textdagger}                                                                    & -                             & 56.40                            & 58.30 & -                                 & -                                  & 56.31                            & 57.83 & -                                 & -                                  \\
        BiLSTM \cite{yao2019docred}\textsuperscript{\textdagger}                                                                  & -                             & 58.20                            & 60.04 & -                                 & -                                  & 57.84                            & 59.93 & -                                 & -                                  \\ \midrule
        GAIN \cite{zeng2020double}\textsuperscript{\textdagger}                                                                   & $\text{BERT}_{base}$          & 71.99                            & 73.49 & -                                 & -                                  & 71.88                            & 73.44 & -                                 & -                                  \\
        ATLOP \cite{zhang2021document}  & $\text{BERT}_{base}$  & 73.35\textsuperscript{\textdagger} & 74.22\textsuperscript{\textdagger} & - & -  & 71.86\textsuperscript{*} & 72.61\textsuperscript{*} & 75.92\textsuperscript{*} & 67.46\textsuperscript{*}                                    \\
        PEMSCL \cite{guo2023towards}    & $\text{BERT}_{base}$  & 73.82 & 74.50 & 78.21 & 69.03 & 73.05 & 74.14 & 77.77 & 68.38                              \\
        KMGRE \cite{jiang2022key}\textsuperscript{\textdagger}   & $\text{BERT}_{base}$  & 73.33 & 74.44 & - & - & 73.39 & 74.46 & - & -                                  \\
        NRCL \cite{zhou2022none}    & $\text{BERT}_{base}$          & 73.74                                  & 74.84 & 77.95                             & 70.12                              & 73.40                                  & 74.48 & 77.53                             & 69.59                              \\
        CAST \cite{tan2023class}\textsuperscript{*}    & $\text{BERT}_{base}$  & -  & - & - & - & 73.32   & 74.67 & \textbf{78.53} & 69.34 \\
        KD-DocRE \cite{tan2022document} & $\text{BERT}_{base}$  & 74.37 & 75.31 & 78.71 & 70.26 & 73.98 & 74.88 & 78.28 & 69.52 \\ \midrule
        \textbf{VaeDiff-DocRE (Ours)}   & $\text{BERT}_{base}$  & \textbf{74.96\small$\pm$0.21}   & \textbf{75.89\small$\pm$0.2} & \textbf{79.18\small$\pm$0.22} & \textbf{71.12\small$\pm$0.38} & \textbf{74.13\small$\pm$0.21} & \textbf{75.07\small$\pm$0.2} & 78.32\small$\pm$0.15 & \textbf{70.06\small$\pm$0.29} \\ \midrule
        JEREX \cite{eberts2021end}     & $\text{RoBERTa}_{Large}$  & 71.59 & 72.68 & - & - & 71.45 & 72.57 & 77.09 & 66.31                                  \\
        ATLOP \cite{zhou2022none}                                   & $\text{RoBERTa}_{Large}$  & 76.79 & 77.46 & - & - & 76.82 & 77.56 & 80.78 & 72.29                              \\
        DocuNET \cite{zhang2021document}                            & $\text{RoBERTa}_{Large}$  & 77.49 & 78.14 & - & - & 77.26 & 77.87 & 81.25 & 73.32 \\
        KD-DocRE \cite{tan2022document}                             & $\text{RoBERTa}_{Large}$  & 77.85 & 78.51 & - & - & 77.60 & 78.28 & 80.85 & 74.31 \\ \midrule
        \textbf{VaeDiff-DocRE (Ours)}   & $\text{RoBERTa}_{Large}$  & \textbf{78.35\small$\pm$0.22}   & \textbf{79.19\small$\pm$0.16} & \textbf{82.0\small$\pm$0.18} & \textbf{75.13\small$\pm$0.35} & \textbf{78.22\small$\pm$0.27} & \textbf{79.03\small$\pm$0.21} & \textbf{81.84\small$\pm$0.21} & \textbf{74.73\small$\pm$0.32} \\ \bottomrule
    \end{tabular}%
    }
    \caption{Experimental result on the development and test set of Re-DocRED. \textbf{Bold} is best performance. \textsuperscript{\textdagger}, \textsuperscript{*} are reported from \citet{jiang2022key} and \citet{tan2023class} respectively. The result on $\text{RoBERTa}_{Large}$ PLM is reported from \citet{tan2022revisiting}. Our results are average of ten runs and conducted statistical significant test following \citet{dror2018hitchhiker}.}
    \label{tab:redocred-perf}
\end{table*}

\begin{table}[!h]
    \resizebox{\columnwidth}{!}{
    \begin{tabular}{lcccc}
    \toprule
    \multirow{2}{*}{\textbf{Method}}        & \multicolumn{2}{c}{\textbf{Dev}}        & \multicolumn{2}{c}{\textbf{Test}}      \\ \cmidrule(l){2-3} \cmidrule(l){4-5}
                                            & Ign F1  & F1                    & Ign F1  & F1                  \\ \midrule
    CNN             & 37.65   & 47.73   & 34.65   & 46.14               \\
    LSTM            & 40.86   & 51.77   & 40.81   & 52.60               \\
    BiLSTM          & 40.46   & 51.92   & 42.03   & 54.47               \\ \midrule
    Context-Aware   & 42.06   & 53.05   & 45.37   & 56.58               \\
    CorefBERT       & 57.18   & 61.42   & 61.71   & 66.59               \\
    GAIN            & 58.63   & 62.55   & 62.37   & 67.57               \\
    SSAN            & 58.62   & 64.49   & 62.58   & 69.39               \\
    ATLOP           & 59.03   & 64.82   & 62.09   & 69.94               \\
    RSMAN           & 60.02   & 65.88   & 63.42   & 70.95               \\
    Correl          & 61.10   & 65.73   & 65.64   & 71.56   \\ \midrule
    \textbf{VaeDiff-DocRE (ours)}           & \textbf{61.51\small$\pm$0.58}   & \textbf{67.77\small$\pm$0.46}         & \textbf{66.44\small$\pm$0.48}        & \textbf{73.07\small$\pm$0.37}      \\ 
    \bottomrule
    \end{tabular}
    }
    \caption{Comparison result on the DWIE dataset. The PLM encoder is BERT-base-cased \cite{devlin2019bert}. Our results are average of ten runs. The results are reported from \citet{han2024document}. \textbf{Bold} is best performance.}
    \label{tab:dwie-perf}
\end{table}

\subsection{Experimental Setup}
\paragraph{Dataset.}
We evaluated our proposed method using two benchmark datasets: Re-DocRED \cite{tan2022revisiting} and DWIE \cite{zaporojets2021dwie}. Re-DocRED is a high-quality, revised version of DocRED \cite{yao2019docred} -- a large-scale document-level RE dataset constructed from Wikipedia and Wikidata, which supplements previously missed false negative triplets. DWIE is a document-level information extraction dataset that includes relation extraction. The statistics for these two datasets are provided in Table \ref{tab:dataset-statistics}.

\paragraph{Metrics.}
Following \citet{yao2019docred}, we use micro F1 (\textbf{F1}) and micro Ignore F1 (\textbf{Ign F1}) as evaluation metrics to assess overall DocRE performance. Ign F1 measures the F1 score while excluding relational triplets shared between the training and test sets. Additionally, we use two F1 scores specific to long-tailed distribution settings proposed by \citet{tan2022revisiting}: \textbf{Freq. F1} for common relations and \textbf{LTail F1} for uncommon relations. For the Re-DocRED datasets, Freq F1 is calculated for the 10 most common relations, while LT F1 covers the remaining 86 relations.

\paragraph{Implementation details.}
We implemented our method using Huggingface's Transformers \cite{wolf2019huggingface} and PyTorch 2.1.2 \cite{paszke2019pytorch}. For the document encoders, we utilized BERT-base-cased \cite{devlin2019bert} and RoBERTa-large \cite{liu2019roberta}. The model was optimized using the AdamW optimizer \cite{loshchilov2018decoupled}. All models were implemented and trained on an NVIDIA H100 GPU. We utilized the public repositories of baseline models to implement our experiments.

In stage 1, we configure the learning rates for the document encoder at $3e-5$ and the classifier head at $1e-4$ for the DocRED and Re-DocRED datasets, while setting $3e-4$ for the DWIE dataset. The training process is conducted over 30 epochs. In stage 2, we set the latent size of the EP-VAE to be 384 and employ a Cosine Annealing schedule \cite{loshchilov2022sgdr} for the learning rate, ranging from $2e-5$ to $2e-4$. For the Diffusion Prior, the diffusion step is set to $T=50$, utilizing a Cosine Noise Scheduler \cite{nichol2021improved}. Our denoising architecture consists of 6 layers with a dimension of 256. During training, the probability $p$ of self-condition and class condition is 0.1, and the hyper-parameter $w$ in Eq. (\ref{eq:diff-inf}) is set to 0.1. The VaeDiff model is trained over 100 epochs. In stage 3, the initial training epochs $n$ is 5 epochs on the Re-DocRED datasets and 10 epochs on the DWIE dataset. We set $m = 2$. The overall training epochs for the DocRE models are 30 for Re-DocRED and 50 for DWIE.

\subsection{Baseline}
\paragraph{Re-DocRED} 
For the Re-DocRED dataset, we compared our model with the following baselines in addressing PP imbalance: (1) \textbf{KD-DocRE} \cite{tan2022document} introduces a loss function based on Focal Loss \cite{lin2017focal}, specifically tailored for the DocRE task under long-tail distribution settings. (2) \textbf{PEMSCL} \cite{guo2023towards} proposes a modified supervised contrastive loss function to address similar challenges in the task. (3) \textbf{CAST} \cite{tan2023class} combines self-training with a pseudo-label re-sampling method to improve performance on infrequent relations.

\paragraph{DWIE} 
For the DWIE dataset, we compared our method with the state-of-the-art (SOTA) method in address the PP imbalance -- \textbf{Correl} \cite{han2024document}. It introduces a module designed to learn the relational correlation by leveraging the semantic distance between common and uncommon relations to enhance performance on the latter.

\begin{table}[!ht]
\centering
\resizebox{\columnwidth}{!}{%
    \begin{tabular}{@{}cccccc@{}}
    \toprule
    \multicolumn{2}{c}{\textbf{Datasets}} & \textbf{\# Doc.} & \textbf{\# Rel.}    & \textbf{\# Ent.} & \textbf{\# Facts} \\ \midrule
    \multirow{3}{*}{Re-DocRED}   & Train  & 3,053            & \multirow{3}{*}{96} & 59,359           & 85,932            \\
                                 & Dev    & 500              &                     & 9,684            & 17,284            \\
                                 & Test   & 500              &                     & 9,779            & 17,448            \\ \midrule
    \multirow{3}{*}{DWIE}        & Train  & 602              & \multirow{3}{*}{65} & 16,494           & 14,403            \\
                                 & Dev    & 98               &                     & 2,785            & 2,624             \\
                                 & Test   & 99               &                     & 2,623            & 2,459             \\ \bottomrule
    \end{tabular}%
}
\caption{Statistics of DWIE and Re-DocRED.}
\label{tab:dataset-statistics}
\end{table}

\subsection{Main Results}
\noindent
We present the performance of our proposed VaeDiff-DocRE method for the DocRE task, benchmarking it against baseline methods in Table \ref{tab:redocred-perf} and Table \ref{tab:dwie-perf}. To validate the significance of our improvements, we performed significance tests following the methodology outlined by \citet{dror2018hitchhiker}, with all $p$-values being less than 0.05. These tables illustrate that our VaeDiff-DocRE framework surpasses state-of-the-art methods, particularly in both overall and long-tail performance measured by the F1 score.

In the Re-DocRED dataset, as presented in Table \ref{tab:redocred-perf}, the VaeDiff-DocRE approach demonstrated superior performance compared to baseline methods, achieving an average F1 score improvement of 0.58 and 0.68 on the development set and 0.19 and 0.75 on the test set, when utilizing BERT and RoBERTa as document encoders, respectively. Furthermore, all PLM-based approaches, including ours, outperformed LLM-based methods by a substantial margin. Regarding the long-tailed distribution, particularly for minority relations, our method outperformed the SOTA method by 0.54 and 0.42 points in F1 score on the test sets. While VaeDiff-DocRE achieved the second-highest performance for majority relations on the test set when using BERT, it achieved the best performance for majority relations with RoBERTa. Consequently, VaeDiff-DocRE secured the highest overall F1 score for both PLMs. Similarly, in the DWIE dataset, as shown in Table \ref{tab:dwie-perf}, VaeDiff-DocRE surpassed the SOTA method by 1.6 and 2.27 points in F1 scores on the development and test sets, respectively. These results underscore the effectiveness of our approach in addressing the challenges posed by long-tail distributions while maintaining strong performance for majority classes, thereby enhancing overall performance in the DocRE task.

\subsection{Ablation Study}
\noindent
We conducted an ablation study on the Re-DocRED dataset to demonstrate the effectiveness of each proposed component, and the results are presented in Table \ref{tab:ablation-study}.

First, we removed the VaeDiff module, and instead added Gaussian noise to the representations of positive entity pairs in the current batch $\mathcal{B}$, referred to as \textit{Gaus Noise}. The results show that simply adding Gaussian noise to the entity pair representations yields better performance compared to the KD-DocRE method. However, this basic augmentation technique performs worse than our method, further highlighting the effectiveness of the proposed approach.

Second, we did not augment new entity pair representations during training, referred to as \textit{No Aug}. This model corresponds to the baseline DocRE model used in stage 1 ($\S$\ref{stage-1}). As shown in the results, this led to a significant drop in performance, with an overall F1 score decrease of 0.69. However, when augmentation was introduced, performance improved for both majority and minority relations. Notably, augmenting new data increased the F1 score for minority relations by nearly 1\% in the LTail F1 score, demonstrating the positive impact of data augmentation in addressing the class-imbalance issue of the DocRE task.

\begin{table}[!h]
    \centering
    \resizebox{\columnwidth}{!}{%
    \begin{tabular}{@{}lcccc@{}}
    \toprule
    \textbf{Model}                                                          & \textbf{Ign F1} & \textbf{F1} & \textbf{Freq F1} & \textbf{LTail F1} \\ \midrule
    VaeDiff-DocRE (Ours)               & \textbf{74.96}  & \textbf{75.89}  & \textbf{79.18}   & \textbf{71.12}    \\
    \ \ \ - \textit{Gaus Noise}   & \underline{74.51}           & \underline{75.44}           & \underline{78.80}            & \underline{70.61}             \\
    \ \ \ - \textit{No Aug}     & 74.23           & 75.20           & 78.74            & 69.97             \\ 
    KD-DocRE                    & 74.37           & 75.31           & 78.71            & 70.26              \\
    \bottomrule
    \end{tabular}%
    }
    \caption{Ablation study on the Re-DocRED development set. The PLM is BERT-base-cased \cite{devlin2019bert}. Second highest performance is \underline{underline}.}
    \label{tab:ablation-study}
\end{table}

\subsection{Analysis of Generated Features}
\begin{figure}[h]
    \centering
    \includegraphics[width=\columnwidth]{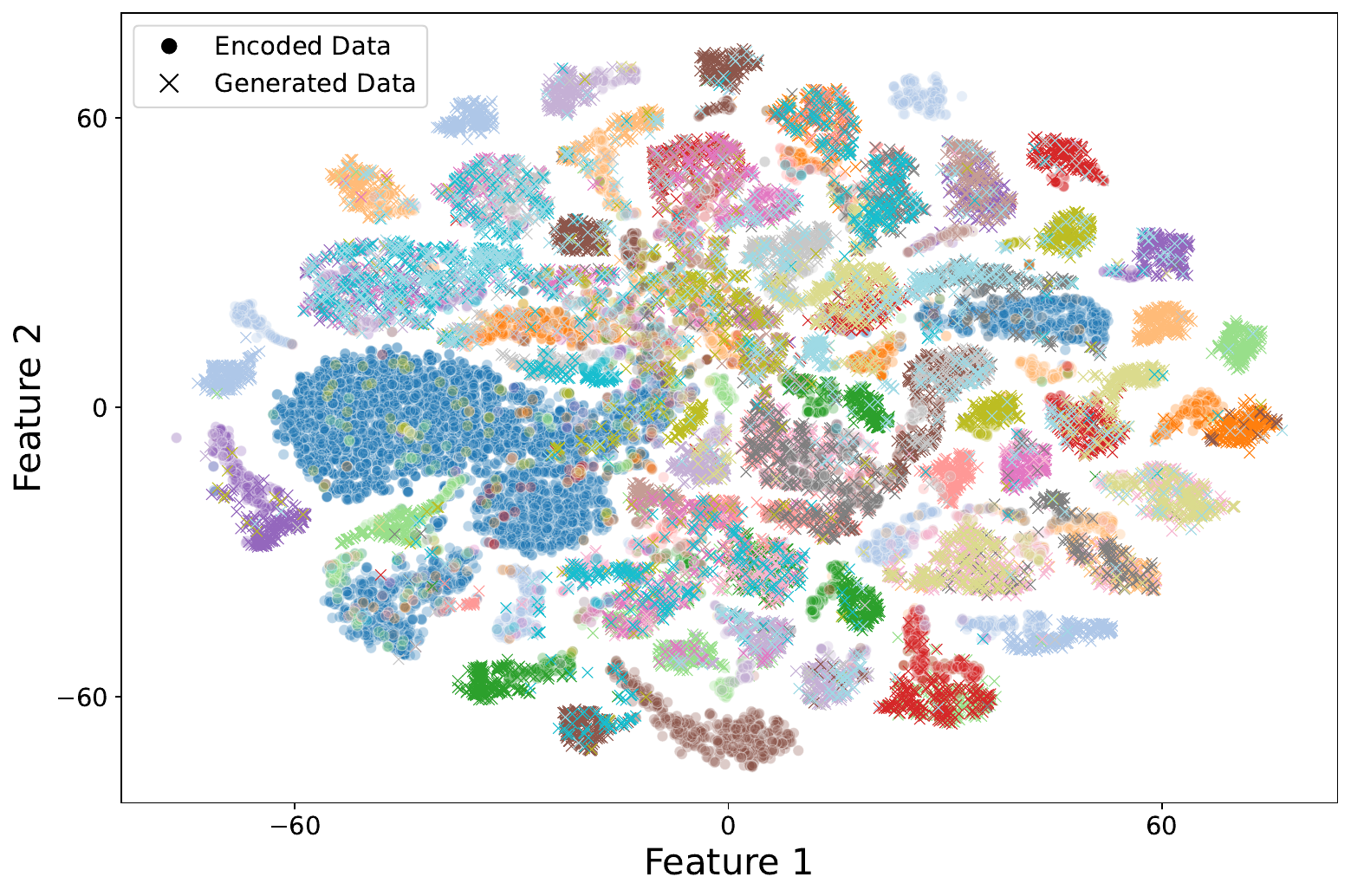}
    \caption{t-SNE visualization of relation-wise distributions of encoded and generated Entity Pair Representations by the trained VaeDiff module ($\S$\ref{sec-vaediff}). Each distribution is depicted in a unique color. \CIRCLE \ represents the encoded actual entity pair representations within the document and $\times$ denotes the representations generated by the module.}
    \label{fig:viz-gen-feats}
\end{figure}

\noindent
In Figure \ref{fig:viz-gen-feats}, we visualize the generated representations of entity pairs expressing minority relations using our VaeDiff approach. By comparing these with the encoded features from the document encoder of the trained DocRE model on the development set of the Re-DocRED dataset, we observe a strong alignment between the generated pair representations and the encoded features. This alignment addresses the lack of high-quality data in minority classes by enriching the true entity pair representations for these relations, effectively enhancing the performance of DocRE models in long-tail distribution settings. However, Figure \ref{fig:viz-gen-feats} also shows that some entity pair representations expressing different relations overlap significantly. This overlap occurs because these relations frequently co-occur in the dataset. For instance, entity pairs expressing the "P580" relation are highly likely to also express the "P582" and "P585" relations in the Re-DocRED dataset.

%% file: sections/related_work.tex
\section{Related Work}
\paragraph{Document-level Relation Extracion (DocRE)}
Early works in Relation Extraction (RE) focused on sentence-level tasks, where context was confined to a single sentence. However, many relational facts require synthesizing information across multiple sentences. Since \citet{yao2019docred} introduced DocRED, a large-scale document-level RE dataset, various methods have been developed for the DocRE task. One approach leverages Graph Neural Networks (GNNs) to transform documents into graph structures, as seen in the work of \citet{nan2020reasoning} and \citet{zeng2020double}. Another approach uses Transformer architectures, with \citet{zhou2021document} and \citet{tan2022document} improving entity pair representation and reasoning through attention mechanisms.

\paragraph{Long-tail Issue in DocRE}
Real-world DocRE datasets often have a skewed distribution, where a few relations dominate most instances. This creates two challenges: an imbalance between negative and positive instances, and an imbalance among the positive instances themselves. The first issue has been widely studied. For instance, \citet{zhou2021document} introduced an adaptive thresholding loss that treats {\tt NA} logits as a learnable threshold, while \citet{zhou2022none} reframed the task as label-ranking, and \citet{guo2023towards} used pairwise ranking to better separate {\tt NA} from positive relations. However, the second issue—imbalance among positive instances—has received less attention. \citet{tan2022document} tackled it by enhancing the loss function from \citet{zhou2021document} to prioritize uncommon relations using Focal Loss \cite{lin2017focal}. Similarly, \citet{guo2023towards} adapted Supervised Contrastive Loss \cite{khosla2020supervised} for the same goal. Yet, while these methods boost performance on rare relations, they often reduce accuracy on common classes, leading to suboptimal overall results.

\paragraph{Variational Autoencoder in RE}  
To mitigate the heavy reliance on labeled data in the RE task, several studies have explored unsupervised approaches. To learn feature-rich representations that are beneficial in scenarios with limited labeled data, \citet{marcheggiani2016discrete} introduced a method leveraging VAEs. In this approach, the VAE encoder is trained as a relation classifier, while the decoder reconstructs the entity pairs from the input sentences based on the encoder's classifications. This setup allows the decoder to provide a supervisory signal to the encoder. Building on this, \citet{yuan2021unsupervised} proposed a method where the relation classification \( r \) is treated as an intermediate variable conditioned on the input sentence \( x \). In their VAE framework, the latent variable \( z \) is sampled based on the joint contributions of \( r \) and \( x \), and the decoder uses \( r \) and \( z \) to reconstruct the entity pairs within the sentence \( x \).

%% file: sections/conclusion.tex
\section{Conclusion}
\noindent
In this paper, we present VaeDiff, a novel data augmentation framework aimed at addressing long-tail distribution challenges in the DocRE task. Our approach utilizes a Variational Autoencoder (VAE) to learn relation-wise distributions and a Diffusion Probabilistic Model to enhance the EP-VAE's latent space for multi-label scenarios. We also propose a hierarchical framework VaeDiff-DocRE to boost DocRE system performance under long-tail conditions by integrating the VaeDiff module. Extensive experiments and ablation studies show that our framework outperforms SOTA methods.

%% file: sections/limitation.tex
\section*{Limitation}
\noindent
The proposed VaeDiff-DocRE framework requires multiple training rounds and inherits the same limitations as the original Diffusion Probabilistic Model framework. As a result, the GPU training time for this framework is longer compared to other models. Additionally, while the framework has been tested in the general domain, its performance in more specialized domains, such as the medical field, has yet to be evaluated.

%% file: sections/acknowledgements.tex
\section*{Acknowledgements}
\noindent
This research is supported by the Australian Research Council under Grant No. DE210100160.

%% file: sections/appendix.tex
\appendix
\section{Background}
\label{sec:appendix}

\subsection{Variational Autoencoder (VAE)}
\label{vae}
In variational autoencoder (VAE), a set of input data is denoted as $\X=(\x_1,...,\x_n)$ where $n$ denotes the number of total input samples. The latent variables are denoted by vector $\z$. The VAE includes a encoder network including network and variational parameters $\bphi$  that produces variational probability model $q_{\bphi}(\z|\x)$ and a decoder network parameterized by $\btheta$ to reconstruct sample $\tilde{\x}=f_{\btheta}(\z)$. The objective of VAE is to maximize the Evidence Lower BOund (ELBO) \cite{kingma2013auto} as follows:
\begin{equation}
    \max_{\btheta, \bphi} \left\{\sum_{i=1}^n \text{ELBO}(\btheta, \bphi; \x_i)\right\}.
\label{eq:VAE}
\end{equation}
where:
\begin{equation}
    \begin{aligned}
        \text{ELBO}(\btheta, \bphi; \x) &:= \EE_{q_{\bphi}(\z|\x)} [\log(p_{\btheta}(\x|\z))] \\ 
        &- \text{KL}(q_{\bphi}(\z|\x) || p(\z))
    \label{eq:ELBO}
    \end{aligned}
\end{equation}

The first term in Eq. (\ref{eq:ELBO}) corresponds to the negative reconstruction loss between input $\x$ and reconstruction $\tilde{\x}$ under Gaussian parameterization of the output.
The second term in Eq. (\ref{eq:ELBO}) refers to the KL divergence between the variational distribution $q_{\bphi}(\z|\x)$ and the prior distribution $p(\z)$. In general, the prior distribution is assumed to be an isotrophic Gaussian distribution.

\subsection{Diffusion Probabilistic Model}
\label{ddpm}
Given an observation of data $\mathbf{x}_0$, diffusion models \cite{ho2020denoising} learn the data distribution $p(\mathbf{x}_0)$ by reversing a diffusion process.
The diffusion (forward) process is a Markov chain that corrupts the sampled data $\mathbf{x}_0$ by gradually adding random noise to it:
\begin{equation}
    q(\mathbf{x}_t|\mathbf{x}_{t-1})
    =\mathcal{N}(
    \sqrt{1-\beta_t}\mathbf{x}_{t-1}, \beta_t\mathbf{I})
\end{equation}
where $\beta_{1:T}$ are the pre-defined noise variances, $\beta_t\in(0, 1)$ at time step $t$.
When $\beta_t\rightarrow T$, the data distribution will be corrupted to $\mathcal{N}(0,\mathbf{I})$.
By defining $\alpha_t=\prod^t_{i=1}(1-\beta_i)$, we can directly get $\mathbf{x}_t$ by adding noise to the input as follows:
\begin{equation}
    q(\mathbf{x}_t|\mathbf{x}_0) = \mathcal{N}(\sqrt{\alpha}_t\mathbf{x}_0, (1-\alpha_t)\mathbf{I})
\end{equation}
where $\alpha_t\in(0,1)$.

Given access to the original data $\mathbf{x}_0$, the backward process can be inverted analytically:
\begin{equation}
    p(\mathbf{x}_{t-1}|\mathbf{x}_t, \mathbf{x}_0)=\mathcal{N}(f_t(\mathbf{x}_t,\mathbf{x}_0), \sigma_t^2\mathbf{I})
\label{eq:reverse-given-data}
\end{equation}
where $\sigma_t$ can be derived from $\beta_t$, $f_t(\mathbf{x}_t,\mathbf{x}_0)$ has a closed form parameterized by $t$ \cite{ho2020denoising}.
However, since the original data $\mathbf{x}_0$ is not available in the actual generation process, (i.e., the response is supposed to be generated), we can not directly use Eq. (\ref{eq:reverse-given-data}) to sample data and thus approximate $f_t(\cdot)$ using a neural network with the parameter $\boldsymbol\psi$, namely \textit{denoising network}.

The training of a denoising diffusion network is defined by minimizing the Kullback-Leibler divergence \( D_\text{KL}\left[q(\x_{0:T}) \ || \ p_\psi(\x_{0:T})\right] \) between the forward trajectory \( q(\x_{0:T}) \) and the backward trajectory \( p_\psi(\x_{0:T}) \). This divergence term can be decomposed as follows:

\begin{equation}
    \label{eq:diff-elbo-decomp}
    \resizebox{.87\linewidth}{!}{
        $\begin{aligned}
            D_\text{KL}\left[q(\x_{0:T}) || p_\psi(\x_{0:T})\right] = \ & \mathbb{E}_{\x_0 \sim q}\left[\log q(\x_0)\right] - \\
            & \mathbb{E}_{\x_{0:T} \sim q}\left[\log \frac{p_\psi(\x_{0:T})}{q(\x_{1:T}|\x_0)}\right]
        \end{aligned}$
     }
\end{equation}

The second term in Eq. (\ref{eq:diff-elbo-decomp}) represents the expectation of an evidence lower bound (ELBO) with respect to \( \x_0 \sim q \):
\begin{equation}
    \label{eq:elbo-diff}
    \resizebox{.87\linewidth}{!}{
        $\text{ELBO} := \mathbb{E}_{\x_{0:T} \sim q}\left[\log \frac{p_\psi(\x_{0:T})}{q(\x_{1:T}|\x_0)}\right] \leq \log p_\psi(\x_0)$
     }
\end{equation}

By leveraging the Gaussian transitions and simplifying further, \citet{ho2020denoising} express the loss function in a more tractable form:
\begin{equation}
    \mathbb{E}_{t, \mathbf{x}_0, \mathbf{x}_t}\left[\frac{1}{2\sigma_t^2} \norm{f_{\boldsymbol\psi}(\mathbf{x}_t, t)-\mathbf{x}_0}^2_2\right]
\label{eq:diff_loss}
\end{equation}
where $t\sim \text{Uniform}(\{1,\cdots,T\})$, $\mathbf{x}_t\sim q(\mathbf{x}_t|\mathbf{x}_0)$.

For inference, we can use the trained denoising network $f_\psi(\mathbf{x}_t, t)$ to build the usable inversion $p_{\psi}(\mathbf{x}_{t-1}|\mathbf{x}_t)\approx p(\mathbf{x}_{t-1}|\mathbf{x}_t,f_{\boldsymbol\psi}(\mathbf{x}_t, t))$ and get new high-quality data by sampling from it iteratively.

\subsection{Baseline DocRE}
\label{baseline-docre}
\noindent
This section introduces the baseline DocRE model used in our framework introduced in Section \ref{stage-1}.

\subsubsection{Representation Learning}
\paragraph{Encoding}
We insert a special token ``\textbf{*}'' before and behind each mention $m_j^i$ of $e_i$ in document $D$ as an entity marker \cite{zhang2017position}. We use a pre-trained language model (PLM) to obtain token embeddings $H \in \mathbb{R}^{L \times d}$ and token-level cross-attention matrix $A \in \mathbb{R}^{L \times L}$ of $D$:
\begin{equation}
    H, A = PLM([w_1, w_2, ..., w_L])
\end{equation}

The representations of an entity $e_i$ and an entity pair $(e_h, e_t)$ are computed as Eq. (\ref{ent-rep}) and Eq. (\ref{pair-rep}) respectively.
\begin{equation}
    \label{ent-rep}
    \mathbf{e}_i = \text{log} \sum_{i=1}^{|\mathcal{M}_{e_i}|}\text{exp}\left(\mathbf{m}_j^i\right)
\end{equation}
\begin{equation}
    \label{pair-rep}
    \mathbf{p}_{h, t} = \text{tanh}\left(\mathbf{W}_p\left[\mathbf{e}_h; \mathbf{e}_t; \mathbf{c}_{h, t}\right] + \mathbf{b}_p\right)
\end{equation}
where $\mathbf{m}_j^i$ is embedding of starting token "*" of each mention $m_j^i$ of $e_i$; $\mathbf{W}_p$, $\mathbf{b}_p$ are trainable parameters; $[ \ ; \ ]$ is the concatenation; $\mathbf{c}_{h, t}$ is the localized context embedding of $(e_h, e_t)$ \cite{zhou2021document}.

\paragraph{Axial Attention}
After obtaining the representation of entity pairs, we incorporate Axial Attention \cite{ho2019axial} proposed in \cite{tan2022document}, which allows us to integrate neighboring axial information, thereby enhancing the performance of DocRE systems.

Given a set of entities \(\mathcal{E}_D = \{e_i\}_{i=1}^n\) in a document \(D\), we construct an \(n \times n\) entity table, where each row and column corresponds to an entity in \(D\). For any entity pair \((e_h, e_t)\), the axial attention mechanism attends to the pair's axial elements, which can be either \((e_h, e_i)\) or \((e_i, e_t)\). Since multi-hop reasoning is crucial in the DocRE task, if \((e_h, e_t)\) expresses a two-hop relation, axial attention enables the model to explore intermediate relations by connecting through a bridge entity \(e_i\). Specifically, axial attention is computed by applying self-attention \cite{vaswani2017attention} along the height axis and the width axis of the entity table. Each computation along these axes is followed by a residual connection. The process for a pair \((e_h, e_t)\) is as follows:
\begin{equation}\small
    \begin{split}
            r^{(h, t)}_{w} &= r^{(h,t)}_{h} +
            \sum_{{p}\in{1..n}}{\text{softmax}_{p}}({q}_{(h, t)}^{T}k_{(h, p)})v_{(h, p)} \\
            r^{(h, t)}_{h} &= \mathbf{p}_{h, t} +
            \sum_{{p}\in{1..n}}{\text{softmax}_{p}}({q}_{(h, t)}^{T}k_{(p, t)})v_{(p, t)}
    \end{split}
\end{equation}
\noindent
where the query \({q}_{(i, j)} = W_{Q}\mathbf{p}_{i, j}\), key \({k}_{(i, j)} = W_{K}\mathbf{p}_{i, j}\), and value \({v}_{(i, j)} = W_{V}\mathbf{p}_{i, j}\) are linear projections of the entity pair representation \(\mathbf{p}\) at the position \((i, j)\). The matrices \(W_{Q}\), \(W_{K}\), \(W_{V}\) are learnable weight parameters.

\subsubsection{Objective Function}
\noindent
To enhance the robustness of the baseline DocRE system against the significant PN imbalance issue, we utilize Pairwise Moving-Threshold Loss combined with Entropy Minimization \cite{guo2023towards} as the primary objective, denoted as $\mathcal{L}^{h,t}_{pmt}$. 

\paragraph{Pairwise Moving-Threshold Loss with Entropy Minimization ($\mathcal{L}^{h,t}_{pmt}$)}
For each entity pair \((e_h, e_t)\), the relation set \(\mathcal{R}\) is divided into a positive relation set \(\mathcal{P}_{h,t}\) and a negative relation set \(\mathcal{N}_{h,t}\). Given the prediction score \(f_r\) for a relation \(r \in \mathcal{R}\) and the prediction score \(f_\eta\) for the {\tt NA} class of \((e_h, e_t)\), the probability that the relation \(C\) of \((e_h, e_t)\), denoted as \(P^r_{h,t}(r)\), is \(r\) given that \(C\) is either \(r\) or the {\tt NA} class is computed as follows:
\begin{equation}
    P^r_{h,t}(r) = \frac{\exp(f_r)}{\exp(f_r) + \exp(f_\eta)}
\end{equation}
Inversely, the probability of $C$ is {\tt NA} class is as:
\begin{equation}
    P^{\text{\tt NA}}_{h,t}(r) \!=\!1\!-\!P^r_{h,t}(r) 
    \!=\! \frac{\exp(f_\eta)}{\exp(f_r)\!+\!\exp(f_\eta)} \label{eq:p_eta}
\end{equation}
Then, the \textit{pairwise moving-threshold} loss $\mathcal{L}^{h,t}_{pmt}$ is computed as follow:
\begin{align}
    \mathcal{L}^{h,t}_{pmt} & = - \log \Big (\!\!\prod_{r\in \mathcal{P}_{h,t}}\!\!P^r_{h,t}(r) \!\!\prod_{r \in \mathcal{N}_{h,t}}\!\! \big (1 - P^r_{h,t}(r) \big) \Big ) \nonumber \\
    &= -\!\! \sum_{r\in \mathcal{P}_{h,t}}\!\!\log P^r_{h,t}(r)-\!\! \sum_{r\in \mathcal{N}_{h,t}} \!\!\log P^{\text{\tt NA}}_{h,t}(r) \nonumber \\
    &= \sum_{r\in \mathcal{P}_{h,t}} \log(1 + \exp(f_\eta - f_r)) \nonumber \\ 
    &+ \sum_{r\in \mathcal{N}_{h,t}} \log(1 + \exp(f_r-f_\eta)).
    \label{eq:ptm} 
\end{align}
To further discriminate the margin in order to reduce the certainty of model prediction about positive and negative relations, \cite{guo2023towards} compute the information entropy for each pairwise probability distribution between relation $r$ and the {\tt NA} class for each \((e_h, e_t)\):
\begin{equation}
    H_{h,t}(r) = - P^r_{h,t}(r) \log P^r_{h,t}(r) - P^{\text{\tt NA}}_{h,t}(r)\log P^{\text{\tt NA}}_{h,t}(r).  \label{eq:entro}
\end{equation}
Finally, the pairwise moving-threshold loss with entropy minimization is as follows:
\begin{align}
    \mathcal{L}^{h,t}_{em} &\!=\! \frac{1}{\gamma_1}\!\sum_{r\in \mathcal{P}_{h,t}}\!H_{h,t}(r) \!+ \frac{1}{\gamma_2}\!\sum_{r\in \mathcal{N}_{h,t}}\!H_{h,t}(r), \label{eq:em}\\
    \mathcal{L}_1 &= \sum_{(h, t) \in \mathcal{B}} \mathcal{L}^{h,t}_{pmt} + \mathcal{L}^{h,t}_{em},
    \label{eq:l1}
\end{align}
where $\mathcal{B}$ refers to a training batch, and $\gamma_1 =\{1, |\mathcal{P}_{h,t}|\}$ and $\gamma_2 = \{1,|\mathcal{N}_{h,t}|\}$ are hyperparameters weighting the effect of entropy minimization.

\paragraph{Supervised Contrastive Learning ($\mathcal{L}^{h,t}_{scl}$)}
We further incorporate the Supervised Contrastive Learning loss \cite{khosla2020supervised} to enhance the performance the DocRE model:
\begin{equation}
    \resizebox{.87\linewidth}{!}{
    $\begin{aligned}
      \mathcal{L}^{h,t}_{scl} = -\log \left\{ \frac{1}{|\mathcal{S}_{h,t}|}\sum_{p\in \mathcal{S}_{h,t}} \frac{\exp(\bm{x}_{h,t} \cdot \bm{x}_p / \tau)}{\sum\limits_{d \in \mathcal{B},d\neq (h,t)}\exp(\bm{x}_{h,t} \cdot \bm{x}_d/\tau)} \right\}, \label{eq:scl}
  \end{aligned}$}
\end{equation}
where $\mathcal{B}$ is the set of entity pairs including $(h,t)$ in the current batch. $\mathcal{S}_{h,t} \subseteq \mathcal{B}$ is the set in which each entity pair $p\!=\!(h',t')$ has at least one common positive relation with $(h,t) \notin \mathcal{S}_{h,t}$, and $p$ is termed 
a \textit{positive} example of $(h,t)$. The \textit{negative} examples of $(h,t) $ are the remaining examples in $\mathcal{B}$. The operator ($\cdot$) refers to the dot product, and $\tau \in \mathbb{R}^+$ is a temperature parameter.

\paragraph{Final Objective Functions ($\mathcal{L}_{\text{DocRE}}$)}
\noindent
Additionally, we apply the Supervised Contrastive Learning loss, denoted as $\mathcal{L}^{h,t}_{scl}$, to emphasize the distinctions between the embeddings of entity pairs that express different relations. Ultimately, the baseline DocRE model aims to minimize the following loss function:
\begin{equation}
    \label{base-loss}
    \mathcal{L}_{\text{DocRE}} = \mathcal{L}^{h,t}_{pmt} + \lambda \times \mathcal{L}^{h,t}_{scl}
\end{equation}
where $\lambda$ is the hyper-paremeter to control the strength of $\mathcal{L}^{h,t}_{scl}$.

\subsection{Derivation of Equation (\ref{eq:loss-vaediff})}
\label{sec:derive-eq-loss-vaediff}
\noindent
First, Eq. (\ref{eq:ELBO}) is reformulated using Eq. (\ref{eq:elbo-decomp}):  
\begin{equation}
    \label{eq:vae-elbow-decomp}
    \resizebox{.87\linewidth}{!}{
        $\begin{aligned}
            \text{ELBO}(\btheta, \bphi, \psi; \p) & = \mathbb{E}_{q_{\bphi}(\z_\p|\p)} [\log(f_{\btheta}(\p|\z_\p))] \\ 
            & + \mathbb{E}_{q_{\bphi}(\z_\p|\p)}\left[-\log(q_{\bphi}(\z_\p|\p))\right] \\ 
            & + \mathbb{E}_{q_{\bphi}(\z_\p|\p)}\left[\log(p_{\psi}(\z_\p))\right]
        \end{aligned}$
    }
\end{equation}  
where \( p_\psi \) denotes the diffusion model, \( p_\theta \) is the VAE decoder, and \( q_\phi \) represents the VAE encoder.

Next, the final term in Eq. (\ref{eq:vae-elbow-decomp}) is expanded using Eq. (\ref{eq:elbo-diff}), yielding the following:  
\begin{equation}
    \label{eq:combine-elbo-vae-diff-1}
    \resizebox{0.85\linewidth}{!}{
    $\begin{aligned}
        \mathbb{E}_{q_{\bphi}}[\log(f_{\btheta}(\p|\z_{\p_0})) - \log(q_{\bphi}(\z_{\p_0}|\p)) + 
        \\ \mathbb{E}_{\z_{\p_{0:T}} \sim q_\phi}\left[\log \frac{p_\psi(\z_{\p_{0:T}})}{q_{\bphi}(\z_{\p_{1:T}}|\z_{\p_0})}\right]
    \end{aligned}$
    }
\end{equation}  
\begin{equation}
    \label{eq:combine-elbo-vae-diff-2}
    \resizebox{0.87\linewidth}{!}{
        $\begin{aligned}
            \leq \ & \mathbb{E}_{q_{\bphi}}[\log(f_{\btheta}(\p|\z_{\p_0})) - \log(q_{\bphi}(\z_{\p_0}|\p)) \\
            & + \log(p_{\psi}(\z_{\p_0}))] \leq \log(f_{\btheta}(\p|\z_{\p_0}))
        \end{aligned}$
    }
\end{equation}  
where \( \z_{\p_0} \) corresponds to the latent space representation generated by the EP-VAE module.

Eqs. (\ref{eq:combine-elbo-vae-diff-1}) and (\ref{eq:combine-elbo-vae-diff-2}) establish a valid ELBO formulation. Maximizing Eq. (\ref{eq:combine-elbo-vae-diff-1}) is equivalent to minimizing the VaeDiff module's objective function. Additionally, maximizing the final term in Eq. (\ref{eq:combine-elbo-vae-diff-1}) aligns with minimizing the Diffusion Prior’s objective function. By substituting the last term of Eq. (\ref{eq:combine-elbo-vae-diff-1}) with the Diffusion Prior loss, we derive Eq. (\ref{eq:loss-vaediff}).

%% file: main.bbl
\begin{thebibliography}{45}
\providecommand{\natexlab}[1]{#1}

\bibitem[{Chen et~al.(2023)Chen, ZHANG, and Hinton}]{chen2023analog}
Ting Chen, Ruixiang ZHANG, and Geoffrey Hinton. 2023.
\newblock \href {https://openreview.net/forum?id=3itjR9QxFw} {Analog bits: Generating discrete data using diffusion models with self-conditioning}.
\newblock In \emph{The Eleventh International Conference on Learning Representations}.

\bibitem[{Devlin et~al.(2019)Devlin, Chang, Lee, and Toutanova}]{devlin2019bert}
Jacob Devlin, Ming-Wei Chang, Kenton Lee, and Kristina Toutanova. 2019.
\newblock {BERT}: Pre-training of deep bidirectional transformers for language understanding.
\newblock In \emph{Proceedings of the 2019 Conference of the North {A}merican Chapter of the Association for Computational Linguistics: Human Language Technologies, Volume 1 (Long and Short Papers)}, pages 4171--4186, Minneapolis, Minnesota. Association for Computational Linguistics.

\bibitem[{Dror et~al.(2018)Dror, Baumer, Shlomov, and Reichart}]{dror2018hitchhiker}
Rotem Dror, Gili Baumer, Segev Shlomov, and Roi Reichart. 2018.
\newblock The hitchhiker’s guide to testing statistical significance in natural language processing.
\newblock In \emph{Proceedings of the 56th annual meeting of the association for computational linguistics (volume 1: Long papers)}, pages 1383--1392.

\bibitem[{Eberts and Ulges(2021)}]{eberts2021end}
Markus Eberts and Adrian Ulges. 2021.
\newblock An end-to-end model for entity-level relation extraction using multi-instance learning.
\newblock In \emph{Proceedings of the 16th Conference of the European Chapter of the Association for Computational Linguistics: Main Volume}, pages 3650--3660.

\bibitem[{Guo et~al.(2023)Guo, Kok, and Bing}]{guo2023towards}
Jia Guo, Stanley Kok, and Lidong Bing. 2023.
\newblock Towards integration of discriminability and robustness for document-level relation extraction.
\newblock In \emph{Proceedings of the 17th Conference of the European Chapter of the Association for Computational Linguistics}, pages 2606--2617.

\bibitem[{Han et~al.(2024)Han, Peng, Wang, Liu, Tiwari, and Wan}]{han2024document}
Ridong Han, Tao Peng, Benyou Wang, Lu~Liu, Prayag Tiwari, and Xiang Wan. 2024.
\newblock Document-level relation extraction with relation correlations.
\newblock \emph{Neural Networks}, 171:14--24.

\bibitem[{Ho et~al.(2020)Ho, Jain, and Abbeel}]{ho2020denoising}
Jonathan Ho, Ajay Jain, and Pieter Abbeel. 2020.
\newblock Denoising diffusion probabilistic models.
\newblock \emph{Advances in neural information processing systems}, 33:6840--6851.

\bibitem[{Ho et~al.(2019)Ho, Kalchbrenner, Weissenborn, and Salimans}]{ho2019axial}
Jonathan Ho, Nal Kalchbrenner, Dirk Weissenborn, and Tim Salimans. 2019.
\newblock Axial attention in multidimensional transformers.
\newblock \emph{arXiv preprint arXiv:1912.12180}.

\bibitem[{Ho and Salimans(2021)}]{ho2021classifierfree}
Jonathan Ho and Tim Salimans. 2021.
\newblock \href {https://openreview.net/forum?id=qw8AKxfYbI} {Classifier-free diffusion guidance}.
\newblock In \emph{NeurIPS 2021 Workshop on Deep Generative Models and Downstream Applications}.

\bibitem[{Ioffe and Szegedy(2015)}]{ioffe2015batch}
Sergey Ioffe and Christian Szegedy. 2015.
\newblock Batch normalization: Accelerating deep network training by reducing internal covariate shift.
\newblock In \emph{Proceedings of the 32nd International Conference on Machine Learning}, volume~37, pages 448--456. PMLR.

\bibitem[{Jiang et~al.(2022)Jiang, Niu, Mo, and Fan}]{jiang2022key}
Feng Jiang, Jianwei Niu, Shasha Mo, and Shengda Fan. 2022.
\newblock Key mention pairs guided document-level relation extraction.
\newblock In \emph{Proceedings of the 29th International Conference on Computational Linguistics}, pages 1904--1914.

\bibitem[{Khosla et~al.(2020)Khosla, Teterwak, Wang, Sarna, Tian, Isola, Maschinot, Liu, and Krishnan}]{khosla2020supervised}
Prannay Khosla, Piotr Teterwak, Chen Wang, Aaron Sarna, Yonglong Tian, Phillip Isola, Aaron Maschinot, Ce~Liu, and Dilip Krishnan. 2020.
\newblock Supervised contrastive learning.
\newblock \emph{Advances in neural information processing systems}, 33:18661--18673.

\bibitem[{Kingma and Welling(2013)}]{kingma2013auto}
Diederik Kingma and Max Welling. 2013.
\newblock Auto-encoding variational bayes.
\newblock \emph{arXiv preprint arXiv:1312.6114}.

\bibitem[{Li et~al.(2023)Li, Jia, and Zheng}]{li2023semi}
Junpeng Li, Zixia Jia, and Zilong Zheng. 2023.
\newblock Semi-automatic data enhancement for document-level relation extraction with distant supervision from large language models.
\newblock In \emph{Proceedings of the 2023 Conference on Empirical Methods in Natural Language Processing}, pages 5495--5505.

\bibitem[{Li et~al.(2024)Li, Chen, Long, and Zhang}]{li2024llm}
Xingzuo Li, Kehai Chen, Yunfei Long, and Min Zhang. 2024.
\newblock Llm with relation classifier for document-level relation extraction.
\newblock \emph{arXiv preprint arXiv:2408.13889}.

\bibitem[{Lin et~al.(2017)Lin, Goyal, Girshick, He, and Doll{\'a}r}]{lin2017focal}
Tsung-Yi Lin, Priya Goyal, Ross Girshick, Kaiming He, and Piotr Doll{\'a}r. 2017.
\newblock Focal loss for dense object detection.
\newblock In \emph{Proceedings of the IEEE international conference on computer vision}, pages 2980--2988.

\bibitem[{Liu et~al.(2019)Liu, Ott, Goyal, Du, Joshi, Chen, Levy, Lewis, Zettlemoyer, and Stoyanov}]{liu2019roberta}
Yinhan Liu, Myle Ott, Naman Goyal, Jingfei Du, Mandar Joshi, Danqi Chen, Omer Levy, Mike Lewis, Luke Zettlemoyer, and Veselin Stoyanov. 2019.
\newblock Roberta: A robustly optimized bert pretraining approach.
\newblock \emph{arXiv preprint arXiv:1907.11692}.

\bibitem[{Loshchilov and Hutter(2019)}]{loshchilov2018decoupled}
Ilya Loshchilov and Frank Hutter. 2019.
\newblock \href {https://openreview.net/forum?id=Bkg6RiCqY7} {Decoupled weight decay regularization}.
\newblock In \emph{International Conference on Learning Representations}.

\bibitem[{Loshchilov and Hutter(2022)}]{loshchilov2022sgdr}
Ilya Loshchilov and Frank Hutter. 2022.
\newblock Sgdr: Stochastic gradient descent with warm restarts.
\newblock In \emph{International Conference on Learning Representations}.

\bibitem[{Marcheggiani and Titov(2016)}]{marcheggiani2016discrete}
Diego Marcheggiani and Ivan Titov. 2016.
\newblock Discrete-state variational autoencoders for joint discovery and factorization of relations.
\newblock \emph{Transactions of the Association for Computational Linguistics}, 4:231--244.

\bibitem[{Nan et~al.(2020)Nan, Guo, Sekuli{\'c}, and Lu}]{nan2020reasoning}
Guoshun Nan, Zhijiang Guo, Ivan Sekuli{\'c}, and Wei Lu. 2020.
\newblock Reasoning with latent structure refinement for document-level relation extraction.
\newblock In \emph{Proceedings of the 58th Annual Meeting of the Association for Computational Linguistics}, pages 1546--1557.

\bibitem[{Nichol and Dhariwal(2021)}]{nichol2021improved}
Alexander~Quinn Nichol and Prafulla Dhariwal. 2021.
\newblock Improved denoising diffusion probabilistic models.
\newblock In \emph{International conference on machine learning}, pages 8162--8171. PMLR.

\bibitem[{Paszke et~al.(2019)Paszke, Gross, Massa, Lerer, Bradbury, Chanan, Killeen, Lin, Gimelshein, Antiga et~al.}]{paszke2019pytorch}
Adam Paszke, Sam Gross, Francisco Massa, Adam Lerer, James Bradbury, Gregory Chanan, Trevor Killeen, Zeming Lin, Natalia Gimelshein, Luca Antiga, et~al. 2019.
\newblock Pytorch: An imperative style, high-performance deep learning library.
\newblock \emph{Advances in neural information processing systems}, 32.

\bibitem[{Sun et~al.(2024)Sun, Huang, Yang, Tong, Zhang, and Poria}]{sun2024consistency}
Qi~Sun, Kun Huang, Xiaocui Yang, Rong Tong, Kun Zhang, and Soujanya Poria. 2024.
\newblock Consistency guided knowledge retrieval and denoising in llms for zero-shot document-level relation triplet extraction.
\newblock In \emph{Proceedings of the ACM on Web Conference 2024}, pages 4407--4416.

\bibitem[{Tan et~al.(2022{\natexlab{a}})Tan, He, Bing, and Ng}]{tan2022document}
Qingyu Tan, Ruidan He, Lidong Bing, and Hwee~Tou Ng. 2022{\natexlab{a}}.
\newblock Document-level relation extraction with adaptive focal loss and knowledge distillation.
\newblock In \emph{Findings of the Association for Computational Linguistics: ACL 2022}, pages 1672--1681.

\bibitem[{Tan et~al.(2023)Tan, Xu, Bing, and Ng}]{tan2023class}
Qingyu Tan, Lu~Xu, Lidong Bing, and Hwee~Tou Ng. 2023.
\newblock Class-adaptive self-training for relation extraction with incompletely annotated training data.
\newblock In \emph{Findings of the Association for Computational Linguistics: ACL 2023}, pages 8630--8643.

\bibitem[{Tan et~al.(2022{\natexlab{b}})Tan, Xu, Bing, Ng, and Aljunied}]{tan2022revisiting}
Qingyu Tan, Lu~Xu, Lidong Bing, Hwee~Tou Ng, and Sharifah~Mahani Aljunied. 2022{\natexlab{b}}.
\newblock Revisiting docred-addressing the false negative problem in relation extraction.
\newblock In \emph{Proceedings of the 2022 Conference on Empirical Methods in Natural Language Processing}, pages 8472--8487.

\bibitem[{Van~der Maaten and Hinton(2008)}]{van2008visualizing}
Laurens Van~der Maaten and Geoffrey Hinton. 2008.
\newblock Visualizing data using t-sne.
\newblock \emph{Journal of machine learning research}, 9(11).

\bibitem[{Vaswani et~al.(2017)Vaswani, Shazeer, Parmar, Uszkoreit, Jones, Gomez, Kaiser, and Polosukhin}]{vaswani2017attention}
Ashish Vaswani, Noam Shazeer, Niki Parmar, Jakob Uszkoreit, Llion Jones, Aidan~N Gomez, \L~ukasz Kaiser, and Illia Polosukhin. 2017.
\newblock Attention is all you need.
\newblock In \emph{Advances in Neural Information Processing Systems}, volume~30.

\bibitem[{Wang et~al.(2023)Wang, Le, Peng, and Chen}]{wang2023adaptive}
Jize Wang, Xinyi Le, Xiaodi Peng, and Cailian Chen. 2023.
\newblock Adaptive hinge balance loss for document-level relation extraction.
\newblock In \emph{Findings of the Association for Computational Linguistics: EMNLP 2023}, pages 3872--3878.

\bibitem[{Wehenkel and Louppe(2021)}]{wehenkel2021diffusion}
Antoine Wehenkel and Gilles Louppe. 2021.
\newblock Diffusion priors in variational autoencoders.
\newblock In \emph{ICML Workshop on Invertible Neural Networks, Normalizing Flows, and Explicit Likelihood Models}.

\bibitem[{Wolf et~al.(2020)Wolf, Debut, Sanh, Chaumond, Delangue, Moi, Cistac, Rault, Louf, Funtowicz, Davison, Shleifer, von Platen, Ma, Jernite, Plu, Xu, Le~Scao, Gugger, Drame, Lhoest, and Rush}]{wolf2019huggingface}
Thomas Wolf, Lysandre Debut, Victor Sanh, Julien Chaumond, Clement Delangue, Anthony Moi, Pierric Cistac, Tim Rault, Remi Louf, Morgan Funtowicz, Joe Davison, Sam Shleifer, Patrick von Platen, Clara Ma, Yacine Jernite, Julien Plu, Canwen Xu, Teven Le~Scao, Sylvain Gugger, Mariama Drame, Quentin Lhoest, and Alexander Rush. 2020.
\newblock Transformers: State-of-the-art natural language processing.
\newblock In \emph{Proceedings of the 2020 Conference on Empirical Methods in Natural Language Processing: System Demonstrations}, pages 38--45. Association for Computational Linguistics.

\bibitem[{Xiong et~al.(2020)Xiong, Yang, He, Zheng, Zheng, Xing, Zhang, Lan, Wang, and Liu}]{xiong2020layer}
Ruibin Xiong, Yunchang Yang, Di~He, Kai Zheng, Shuxin Zheng, Chen Xing, Huishuai Zhang, Yanyan Lan, Liwei Wang, and Tieyan Liu. 2020.
\newblock On layer normalization in the transformer architecture.
\newblock In \emph{International Conference on Machine Learning}, pages 10524--10533. PMLR.

\bibitem[{Xu et~al.(2016)Xu, Reddy, Feng, Huang, and Zhao}]{xu2016question}
Kun Xu, Siva Reddy, Yansong Feng, Songfang Huang, and Dongyan Zhao. 2016.
\newblock Question answering on freebase via relation extraction and textual evidence.
\newblock In \emph{Proceedings of the 54th Annual Meeting of the Association for Computational Linguistics (Volume 1: Long Papers)}, pages 2326--2336.

\bibitem[{Xu et~al.(2021)Xu, Chen, and Zhao}]{xu2021document}
Wang Xu, Kehai Chen, and Tiejun Zhao. 2021.
\newblock Document-level relation extraction with reconstruction.
\newblock In \emph{Proceedings of the AAAI Conference on Artificial Intelligence}, volume~35, pages 14167--14175.

\bibitem[{Xue et~al.(2024)Xue, Zhang, Dong, and Tang}]{xue2024autore}
Lilong Xue, Dan Zhang, Yuxiao Dong, and Jie Tang. 2024.
\newblock Autore: document-level relation extraction with large language models.
\newblock In \emph{Proceedings of the 62nd Annual Meeting of the Association for Computational Linguistics (Volume 3: System Demonstrations)}, pages 211--220.

\bibitem[{Yao et~al.(2019)Yao, Ye, Li, Han, Lin, Liu, Liu, Huang, Zhou, and Sun}]{yao2019docred}
Yuan Yao, Deming Ye, Peng Li, Xu~Han, Yankai Lin, Zhenghao Liu, Zhiyuan Liu, Lixin Huang, Jie Zhou, and Maosong Sun. 2019.
\newblock Docred: A large-scale document-level relation extraction dataset.
\newblock In \emph{Proceedings of the 57th Annual Meeting of the Association for Computational Linguistics}, pages 764--777.

\bibitem[{Yuan and Eldardiry(2021)}]{yuan2021unsupervised}
Chenhan Yuan and Hoda Eldardiry. 2021.
\newblock Unsupervised relation extraction: A variational autoencoder approach.
\newblock In \emph{Proceedings of the 2021 Conference on Empirical Methods in Natural Language Processing}, pages 1929--1938.

\bibitem[{Zaporojets et~al.(2021)Zaporojets, Deleu, Develder, and Demeester}]{zaporojets2021dwie}
Klim Zaporojets, Johannes Deleu, Chris Develder, and Thomas Demeester. 2021.
\newblock Dwie: An entity-centric dataset for multi-task document-level information extraction.
\newblock \emph{Information Processing \& Management}, 58(4):102563.

\bibitem[{Zeng et~al.(2020)Zeng, Xu, Chang, and Li}]{zeng2020double}
Shuang Zeng, Runxin Xu, Baobao Chang, and Lei Li. 2020.
\newblock Double graph based reasoning for document-level relation extraction.
\newblock In \emph{Proceedings of the 2020 Conference on Empirical Methods in Natural Language Processing (EMNLP)}, pages 1630--1640.

\bibitem[{Zhang et~al.(2021)Zhang, Chen, Xie, Deng, Tan, Chen, Huang, Si, and Chen}]{zhang2021document}
Ningyu Zhang, Xiang Chen, Xin Xie, Shumin Deng, Chuanqi Tan, Mosha Chen, Fei Huang, Luo Si, and Huajun Chen. 2021.
\newblock Document-level relation extraction as semantic segmentation.
\newblock In \emph{Proceedings of the Thirtieth International Joint Conference on Artificial Intelligence}. International Joint Conferences on Artificial Intelligence Organization.

\bibitem[{Zhang et~al.(2023)Zhang, Kang, Hooi, Yan, and Feng}]{zhang2023deep}
Yifan Zhang, Bingyi Kang, Bryan Hooi, Shuicheng Yan, and Jiashi Feng. 2023.
\newblock Deep long-tailed learning: A survey.
\newblock \emph{IEEE Transactions on Pattern Analysis and Machine Intelligence}, 45(9):10795--10816.

\bibitem[{Zhang et~al.(2017)Zhang, Zhong, Chen, Angeli, and Manning}]{zhang2017position}
Yuhao Zhang, Victor Zhong, Danqi Chen, Gabor Angeli, and Christopher~D Manning. 2017.
\newblock Position-aware attention and supervised data improve slot filling.
\newblock In \emph{Conference on Empirical Methods in Natural Language Processing}.

\bibitem[{Zhou et~al.(2021)Zhou, Huang, Ma, and Huang}]{zhou2021document}
Wenxuan Zhou, Kevin Huang, Tengyu Ma, and Jing Huang. 2021.
\newblock Document-level relation extraction with adaptive thresholding and localized context pooling.
\newblock In \emph{Proceedings of the AAAI conference on artificial intelligence}, volume~35, pages 14612--14620.

\bibitem[{Zhou and Lee(2022)}]{zhou2022none}
Yang Zhou and Wee~Sun Lee. 2022.
\newblock None class ranking loss for document-level relation extraction.
\newblock In \emph{Proceedings of the Thirty-First International Joint Conference on Artificial Intelligence, {IJCAI-22}}, pages 4538--4544. International Joint Conferences on Artificial Intelligence Organization.

\end{thebibliography}
